\documentclass[a4paper,fleqn]{cas-dc}

\usepackage[authoryear]{natbib}

\usepackage{amsmath,amssymb,amsfonts}
\usepackage{booktabs}
\usepackage{multirow}
\usepackage{pifont}
\usepackage{xcolor}
\usepackage{colortbl}
\usepackage{tikz}
\usepackage{textcomp}
\usepackage{fontawesome5}

\definecolor{metablue}{RGB}{0,100,224}
\hypersetup{colorlinks,linkcolor=metablue,citecolor=metablue,urlcolor=metablue}

\usepackage[capitalize]{cleveref}
\crefname{figure}{Fig.}{Figs.}
\Crefname{figure}{Fig.}{Figs.}
\crefname{table}{Table}{Tables}
\crefname{section}{Section}{Sections}
\crefname{subsection}{Section}{Sections}
\crefname{equation}{Eq.}{Eqs.}
\creflabelformat{equation}{(#2#1#3)}

\newcommand{\tight}{\looseness=-1\relax}

\newcommand{\tablesetup}{\footnotesize\setlength{\tabcolsep}{2pt}\renewcommand{\arraystretch}{1.1}}

\newcommand{\minisection}[1]{\par\noindent{\bfseries #1}\,}
\definecolor{c1}{HTML}{D6ECF0}  
\newcommand{\ourmodel}{STREAM}
\newcommand{\circlednum}[1]{%
    \tikz[baseline=(X.base)]{\node[draw,circle,inner sep=.3pt,font=\scriptsize] (X) {#1};}%
}

\definecolor{todored}{rgb}{0.8,0,0}

\begin{document}
\let\WriteBookmarks\relax
\def\floatpagepagefraction{1}
\def\textpagefraction{.001}

\shorttitle{STREAM for Video Polyp Segmentation}
\shortauthors{T. Wang et al.}

\title[mode=title]{Reliable Recursive Memory for Causal Multi-scale Video Polyp Segmentation}

\author[1,2]{Tong Wang}[orcid=0000-0001-5141-9360]
\ead{tongwangnj@qq.com}
\author[1]{Wenbo Peng}[orcid=0009-0005-0414-1788]
\author[1]{Yaolei Qi}[orcid=0000-0002-8531-7386]
\author[3]{Yuting He}[orcid=0000-0003-0878-8915]
\author[2]{Yutong Xie}[orcid=0000-0002-6644-1250]
\cormark[1]
\ead{yutong.xie678@gmail.com}
\author[1]{Guanyu Yang}[orcid=0000-0003-3704-1722]
\cormark[1]
\ead{yang.list@seu.edu.cn}

\affiliation[1]{organization={Key Laboratory of New Generation Artificial Intelligence Technology and Its Interdisciplinary Applications (Southeast University), Ministry of Education},
            addressline={No.~2 Sipailou},
            city={Nanjing},
            postcode={210096},
            state={Jiangsu},
            country={China}}
\affiliation[2]{organization={Mohamed bin Zayed University of Artificial Intelligence (MBZUAI)},
            addressline={Masdar City},
            city={Abu Dhabi},
            country={United Arab Emirates}}
\affiliation[3]{organization={School of Medicine, Case Western Reserve University},
            addressline={10900 Euclid Avenue},
            city={Cleveland},
            postcode={44106},
            state={OH},
            country={USA}}

\cortext[1]{Corresponding authors.}

\begin{abstract}
Video polyp segmentation (VPS) is essential for computer-aided
colonoscopy, as it localizes and tracks polyps frame by frame. However,
VPS remains challenging because polyps resemble the mucosa and change
quickly in scale, which calls for history beyond a short clip, and
because blur, specular highlights, and occlusion often degrade only part
of a frame. Existing methods keep history as whole frames, so their cost
grows with the history length, and a partly degraded frame carries its
degraded content to later predictions. To address these issues, we
propose \ourmodel{}, a streaming framework for VPS that extends the
causal multi-scale aggregation (CMA) of our conference work from a fixed
clip to an online stream. \ourmodel{} introduces a Recursive Slot Memory
(RSM) that summarizes multi-scale features and soft foreground and
background cues into a constant number of slots, which the current
features read and pass to CMA, supplying long-range history at a
constant cost. Furthermore, we design Reliability-Aware Slot Writing
(RASW), which estimates a pixel-level quality map from the current
frame, supervised by its local error, and uses it to control which
locations are written into RSM and how much each slot is updated.
Experiments on SUN-SEG and CVC-ClinicDB demonstrate that
\ourmodel{} achieves state-of-the-art performance, improving the Dice of the
conference model from 81.3\% to 85.7\% on the hardest subset while
running at 48 FPS with a constant state of 0.32~MB. Code will be
released at \url{https://github.com/wangtong627/STREAM}.
\end{abstract}


\begin{keywords}
Video polyp segmentation \sep Colonoscopy \sep Recursive memory \sep Reliability-aware memory \sep Streaming inference \sep Temporal modeling
\end{keywords}

\maketitle

\section{Introduction}
\label{sec:introduction}
Colorectal cancer is one of the most common cancers
worldwide~\citep{bray2024globocan} and usually develops from
polyps~\citep{biller2021crc}. Although colonoscopy is the standard
screening tool, many polyps are still missed~\citep{jiang2023miss},
necessitating accurate automatic polyp
segmentation~\citep{bernal2017comparative}. Video polyp segmentation
(VPS)~\citep{ji2021pnsnet,sunseg,hu2024sali,chen2025stddnet,cmsa}
segments the polyp in every frame of a colonoscopy video, providing
localization and tracking cues for colonoscopy.

\begin{figure}[pos=!t]
\centering
\includegraphics[width=\columnwidth]{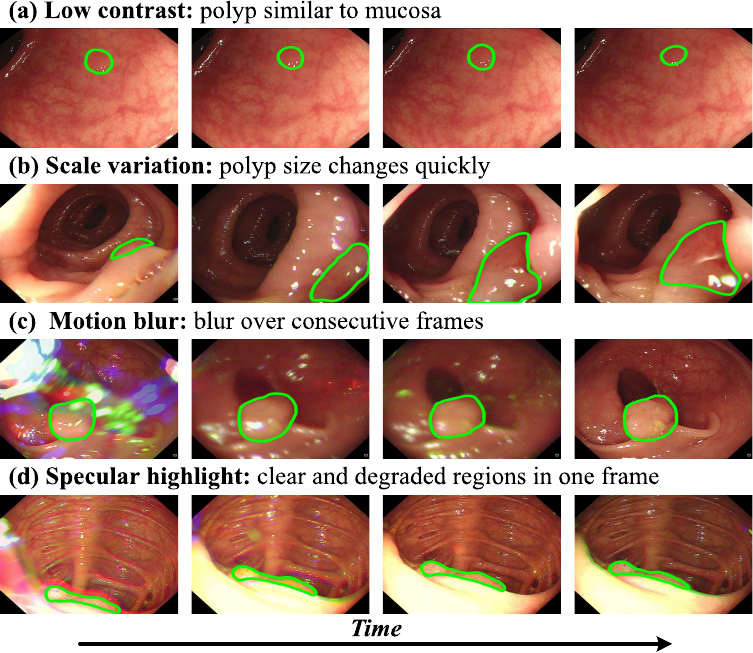}
\caption{\textbf{Challenges in VPS.} \textbf{Challenge 1} (long-range
history dependence): (a) low contrast and (b) fast scale change need
earlier clear frames. \textbf{Challenge 2} (spatially non-uniform
degradation): (c) motion blur and (d) a partial specular
highlight. Green lines mark polyps. Clips (a)--(d):
\texttt{case6\_2}, \texttt{case95\_1}, \texttt{case25\_2}, and \texttt{case51\_2}.}
\label{fig:challenge}
\end{figure}

Compared with image-based
methods~\citep{fan2020pranet,wei2021sanet,wu2022polypsegplus,liu2024boundary},
VPS can use earlier frames to compensate for an ambiguous or degraded
current frame, yet exploiting this context still faces two challenges
(\cref{fig:challenge}). \textbf{(1) Long-range history dependence:}
polyps resemble the mucosa (\cref{fig:challenge}(a)) and change quickly
in scale and viewpoint~\citep{cai2024vanet} (\cref{fig:challenge}(b)), so nearby frames often lack a clear
view, and the useful context lies in earlier clear frames far back in
the video. \textbf{(2) Spatially non-uniform degradation:} artifacts
are common in endoscopy~\citep{ali2021endocv,ali2021quality}; motion blur
can last for several frames (\cref{fig:challenge}(c)), while a highlight
or an instrument often covers only part of a frame
(\cref{fig:challenge}(d)), so the reliability of context varies across
both frames and locations.

\begin{figure}[pos=!t]
\centering
\includegraphics[width=\columnwidth]{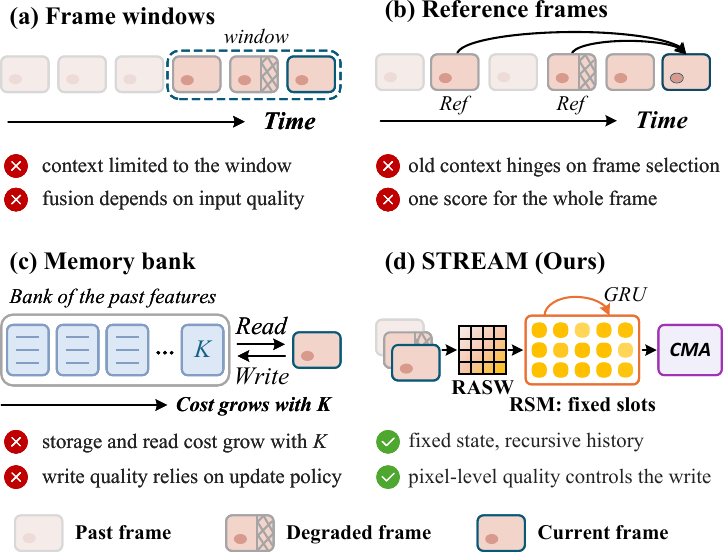}
\caption{\textbf{How VPS methods use history.} (a) Frame windows and
(b) reference frames (e.g., CMSA-Net) keep a few whole frames; (c) memory
banks store past features at a cost that grows with capacity.
(d) \ourmodel{} summarizes history in fixed recursive slots (RSM) and
controls their writing with a pixel-level quality map (RASW).}
\label{fig:paradigm}
\end{figure}

Both challenges concern how history is kept: it should reach earlier
clear frames, and it should exclude degraded content without discarding
the clear content of the same frame. Existing VPS methods keep history
as \textbf{(1) a short frame
window}~\citep{ji2021pnsnet,chen2025stddnet}, \textbf{(2) a few reference
frames}~\citep{sunseg,chen2025mast,cmsa}, or \textbf{(3) a memory bank
of past features}~\citep{hu2024sali} (\cref{fig:paradigm}). Our conference
work CMSA-Net~\citep{cmsa} belongs to the second group: its causal
multi-scale aggregation (CMA) lets the current frame attend to earlier
frames at several scales, and its dynamic multi-source reference (DMR)
keeps two reference frames chosen by frame-level scores. These designs meet neither requirement fully, leaving two limitations. \textbf{(1)~Whole-frame history
(\cref{fig:paradigm}(a)--(c)):} history is kept as whole frames, which
ties temporal coverage to cost. Windows and references, including the
two references of CMSA-Net, fix the cost by keeping a few frames, so an
earlier clear frame (Challenge~1) survives only if it wins a frame-level
selection; memory banks keep more frames at a storage and retrieval cost that grows with their number.
\textbf{(2)~Frame-level write control:} history is updated by keeping or
dropping whole frames by a frame-level score. Under
non-uniform degradation (Challenge~2), a partly degraded frame either
writes its degraded regions into the history, where they mislead later
predictions, or loses its clear ones. Our observations on a CMSA-derived
model (\cref{sec:observation}) show the need for compact long-span history and spatially selective updates.

To address these limitations, we propose \textbf{\ourmodel{}}, a
streaming framework that keeps CMA as its core and runs it online on the
current frame, the cached previous frame, and a memory readout
(\cref{fig:paradigm}(d)). (1)~To overcome whole-frame history, the
\textbf{Recursive Slot Memory (RSM)} summarizes multi-scale features and
soft foreground/background cues into a constant number of GRU-updated
slots, which the current features read and pass to CMA. CMA thus
receives context from earlier clear frames at a cost independent of the
video length. (2)~To replace frame-level write control,
\textbf{Reliability-Aware Slot Writing (RASW)} estimates a pixel-level
quality map from the current frame, supervised by its local error. The map weights
which locations form the written content and gates how much each slot
is updated, so degraded locations contribute less while the clear ones
of the same frame are still written. On SUN-SEG, \ourmodel{} improves
the Dice of CMSA-Net from 81.3\% to 85.7\% on Hard-Unseen with a
constant state of 0.32~MB, regardless of the video length.

Our main contributions are summarized as follows:
\begin{itemize}
\item \textbf{Streaming CMA with reliable history.} Guided by two
observations, we extend CMA from a fixed clip to an online
stream, replacing the window and DMR references with a recursive memory
whose writes are gated by pixel-level reliability.
\item \textbf{Recursive Slot Memory.} RSM serves as the history source
of CMA; a GRU summarizes multi-scale features and soft
foreground/background cues into fixed slots, providing long-range
context at a constant cost.
\item \textbf{Reliability-Aware Slot Writing.} RASW keeps RSM reliable;
a pixel-level quality map supervised by local error weights written
content and gates slot updates.
\item \textbf{Comprehensive validation.} Experiments on SUN-SEG and
CVC-ClinicDB show state-of-the-art performance, improving CMSA-Net by
4.4\% Dice on Hard-Unseen at 48 FPS with a constant 0.32~MB state.
\end{itemize}

A preliminary version of this work was published at MICCAI
2026~\citep{cmsa}. We extend it in four aspects. First, we analyze the
history used by CMA through two motivating observations
(\cref{sec:observation}). Second, we reformulate CMA as a streaming
model with one encoder pass per frame (\cref{sec:cma}). Third, we
replace the discrete references of DMR with RSM (\cref{sec:slot}) and
control what enters it with RASW (\cref{sec:quality}). Finally, we add a second dataset,
state-size comparisons, scenario-based analysis, and ablations
(\cref{sec:experiments}).

\section{Related Work}
\label{sec:related}

\minisection{Video Polyp Segmentation.}
Image-based methods segment each frame independently with reverse
attention~\citep{fan2020pranet}, context selection and shallow
attention~\citep{zhang2020acsnet,wei2021sanet}, lightweight
designs~\citep{wu2022polypsegplus}, transformers~\citep{xiao2024ctnet},
boundary cues~\citep{liu2024boundary}, multi-scale
aggregation~\citep{wang2024sepnet}, state-space
models~\citep{xu2024polypmamba,fu2025polymamba}, frequency-aware
hierarchical learning~\citep{wang2025dshln}, viewpoint
awareness~\citep{cai2024vanet}, uncertainty modeling~\citep{du2025umnet},
gaze-based prompts~\citep{ren2025ppsm}, and confidence-aware
resampling~\citep{guo2022car}. Video polyp segmentation
(VPS) methods further use other frames. PNS-Net~\citep{ji2021pnsnet},
PNS+~\citep{sunseg}, MAST~\citep{chen2025mast}, VP-SAM~\citep{fang2024vpsam},
and STDDNet~\citep{chen2025stddnet} use consecutive frames or a few anchor
or reference frames, and later works model inter-frame cues with
semantic calibration~\citep{wu2021precise}, region
linking~\citep{fang2024embedding}, diffusion~\citep{lu2024diffvps},
deformable alignment~\citep{zhao2025deformable}, and frequency
features~\citep{zhang2026wavepolyp,he2026hfsti}, or reduce
annotation cost with semi-supervision~\citep{zhao2022sstan,chen2026vpsentry}
and synthetic image--mask pairs~\citep{wang2026lamp}. Video polyp
detectors similarly exploit 3D convolutions~\citep{gonzalez2022hybrid},
temporal queues~\citep{wang2025tsdetector}, and spatiotemporal
memory~\citep{hu2026causal}. However, most of these methods keep history
as a short window or a few whole frames. SALI~\citep{hu2024sali}
stores past features in a fixed-length memory bank, and our
CMSA-Net~\citep{cmsa} aggregates past frames at multiple scales (CMA) and
keeps two selected reference frames (DMR), so its history is still a
few whole frames. Closest to our work, \citet{xiong2026qapa} keep two recurrent foreground/background
slots, gate their update with one frame-level reliability value, and use
them to query a long-term prototype bank. \ourmodel{} instead keeps 32 slots that multi-scale features read
for CMA, and RASW controls both content pooling and per-slot updates
with a pixel-level quality map.

\minisection{Memory in Video Object Segmentation.}
STM~\citep{oh2019stm} stores past frames as key--value memory that grows
with the video. AFB-URR~\citep{liang2020afb} and XMem~\citep{cheng2022xmem}
merge or compress this memory, RMem~\citep{zhou2024rmem} restricts the
bank size, Cutie~\citep{cheng2024cutie} adds an object-level
memory, and SAM2~\citep{ravi2024sam2}, which extends
SAM~\citep{kirillov2023sam} to videos, keeps recent and prompted
frames. QDMN~\citep{liu2022qdmn}, SAMURAI~\citep{yang2024samurai},
and \citet{feng2025srr} select the frames to store by predicted
quality or motion cues, and slot
attention~\citep{locatello2020slot,lee2024gsa} provides a fixed number of
object-centric slots. Unlike these frame-based memories, \ourmodel{} updates a
fixed set of slots recursively, so its state stays constant, and it controls reliability within each update instead of per
frame.

\minisection{Reliability and Quality Estimation.}
Uncertainty estimation~\citep{kendall2017uncertainty} and mask quality
estimation~\citep{huang2019msrcnn} predict where or how much a
segmentation is wrong. In polyp segmentation, uncertainty~\citep{du2025umnet}
and pixel-level confidence~\citep{guo2022car} indicate the reliability of
the output or reweight training. ARTEMIS~\citep{artemis} uses temporal reliability
to select references and down-weight noisy supervision for VPS, and
memory methods use quality to choose frames~\citep{liu2022qdmn} or gate
updates~\citep{xiong2026qapa}, mostly with a frame-level score. In
endoscopy, artifacts such as motion blur and specular reflections are
detected and restored as a preprocessing step~\citep{ali2021quality},
which repairs the input but does not decide what a temporal memory
should keep. RASW instead predicts a pixel-level quality map from the current frame,
supervised by its local error, to control both which content enters the
memory and how much each slot is updated.

\section{Method}
\subsection{Preliminaries: CMSA-Net}
\label{sec:prelim}
\ourmodel{} builds on our conference model CMSA-Net~\citep{cmsa}, which
segments the last frame of a short clip with two components, causal
multi-scale aggregation (CMA) and dynamic multi-source reference (DMR),
summarized below.

\minisection{Encoder and decoder.} Each frame $I_t\in\mathbb{R}^{3\times H\times W}$ is encoded by a backbone
and receptive field blocks (RFB) into three-scale features
$x_s^t\in\mathbb{R}^{C\times H_s\times W_s}$, $s\in\{0,1,2\}$, from fine to
coarse, with $H_s=H/2^{s+3}$ and $W_s=W/2^{s+3}$. A two-stage decoder fuses the enhanced
feature with the finer scales to predict the segmentation mask.

\minisection{Causal multi-scale aggregation.} CMA enhances the coarse feature
of a frame with context from other frames at all scales. Each scale is
resized to the grid of $x_2^t$ and aligned by convolutions,
$a_s^t=\mathrm{Align}_s(x_s^t)\in\mathbb{R}^{C\times H_2\times W_2}$. The aligned coarse feature $a_2^t$
forms the query, and the aligned features of all scales of the visible
frames form the keys and values; the attended context is added back to
$a_2^t$ and refined by a feed-forward network. CMA is \emph{causal}: a
frame sees only references and earlier frames. A clip has six
frames: two references, three adjacent frames, and the current
frame.

\minisection{Dynamic multi-source reference.} \tight DMR maintains the reference
frames during inference. It pools the CMA feature of each frame with the
predicted probability map into foreground and background prototypes.
A semantic score adds their separability (one minus cosine similarity)
to the similarity of the foreground prototype with that of the current
frame, and a confidence score adds one minus the mean prediction
entropy to the same similarity. One reference is kept per score and is
replaced only by a later frame with a higher score, with a cooldown of
five and one frames, respectively.

\minisection{From CMSA-Net to \ourmodel{}.} Two properties of this design matter
here. The history of CMA is a short window plus two whole frames, so
older context survives only if DMR keeps it; and DMR decides with one
score per frame, so a frame enters or leaves the history as a whole.

\subsection{Motivating Observations}
\label{sec:observation}
We examine history storage and updates with a streaming CMSA-derived
diagnostic model.
\minisection{Setup.} SUN-SEG training cases are split into 70\% for
training and 30\% for diagnosis, without benchmark test sets.
Results average three seeds, with one model per seed shared by all
policies and paired inputs. Synthetic corruption isolates memory behavior;
Observation~2 uses oracle masks. We insert blur, highlight, or occlusion intervals, change only the history policy, and report Dice within them.

\begin{figure}[pos=!t]
\centering
\includegraphics[width=\columnwidth]{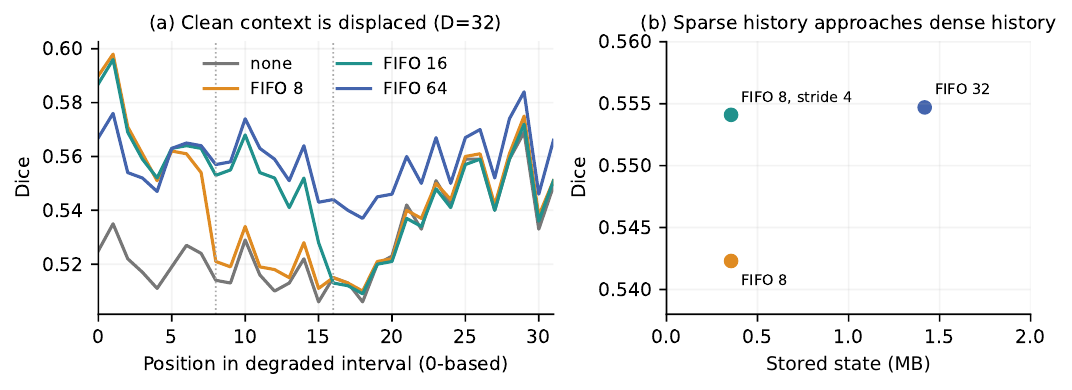}
\caption{\tight \textbf{Observation 1.} Synthetic degradation ($D=32$).
(a) Short FIFO banks lose clean context as degraded entries replace it.
(b) Sparse history nears dense storage at one quarter of the
state.}
\label{fig:obs1}
\end{figure}

\begin{table}[pos=!t]
\centering
\caption{Observation 1: accuracy during $D=32$ synthetic
intervals and storage cost. Dice: mean over three seeds; latency:
median per-frame time on an NVIDIA A100; state in MB.}
\label{tab:obs1}
\tablesetup
\begin{tabular*}{\linewidth}{@{\extracolsep{\fill}}lccc@{}}
\toprule
History & Dice & Latency (ms) & State (MB) \\
\midrule
\circlednum{1} FIFO8 & 0.5423 & 20.41 & 0.354 \\
\circlednum{2} FIFO8, stride 4 & 0.5541 & 20.39 & 0.354 \\
\circlednum{3} FIFO32 & 0.5547 & 20.76 & 1.418 \\
\circlednum{4} FIFO64 & \textbf{0.5580} & 21.13 & 2.836 \\
\circlednum{5} FIFO256 & 0.5509 & 27.18 & 11.344 \\
\bottomrule
\end{tabular*}
\end{table}

\begin{figure}[pos=!t]
\centering
\includegraphics[width=\columnwidth]{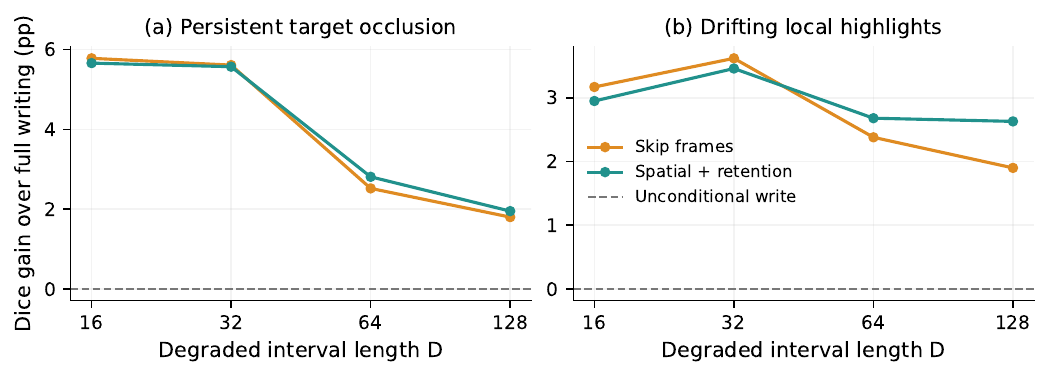}
\caption{\textbf{Observation 2.} Gain over unconditional writing
($K=8$, three-seed means, oracle masks). (a) Both selection
rules help under persistent occlusion. (b) Spatial selection retains
usable updates under drifting highlights. Each gain compares paired
inputs; $D$ settings contain different intervals.}
\label{fig:obs2}
\end{figure}

\begin{table}[pos=!t]
\centering
\caption{Observation 2: Dice during synthetic degradation
($K=8$, three-seed means). Spatial selection uses oracle masks.
Comparisons are within columns.}
\label{tab:obs2}
\tablesetup
\begin{tabular*}{\linewidth}{@{\extracolsep{\fill}}lccc@{}}
\toprule
Write rule & Follow & Scatter & Sweep \\
 & ($D$=32) & ($D$=128) & ($D$=128) \\
\midrule
\circlednum{1} Full write & 0.2709 & 0.4182 & \textbf{0.6359} \\
\circlednum{2} Skip frame & \textbf{0.3270} & 0.4372 & 0.6099 \\
\circlednum{3} Spatial + retain & 0.3266 & \textbf{0.4445} & 0.6312 \\
\circlednum{4} Random + retain & 0.2718 & 0.4215 & 0.6353 \\
\circlednum{5} Spatial, no retain & 0.2712 & 0.4222 & 0.6260 \\
\bottomrule
\end{tabular*}
\end{table}

\begin{figure*}[pos=!t]
\centering
\includegraphics[width=\textwidth]{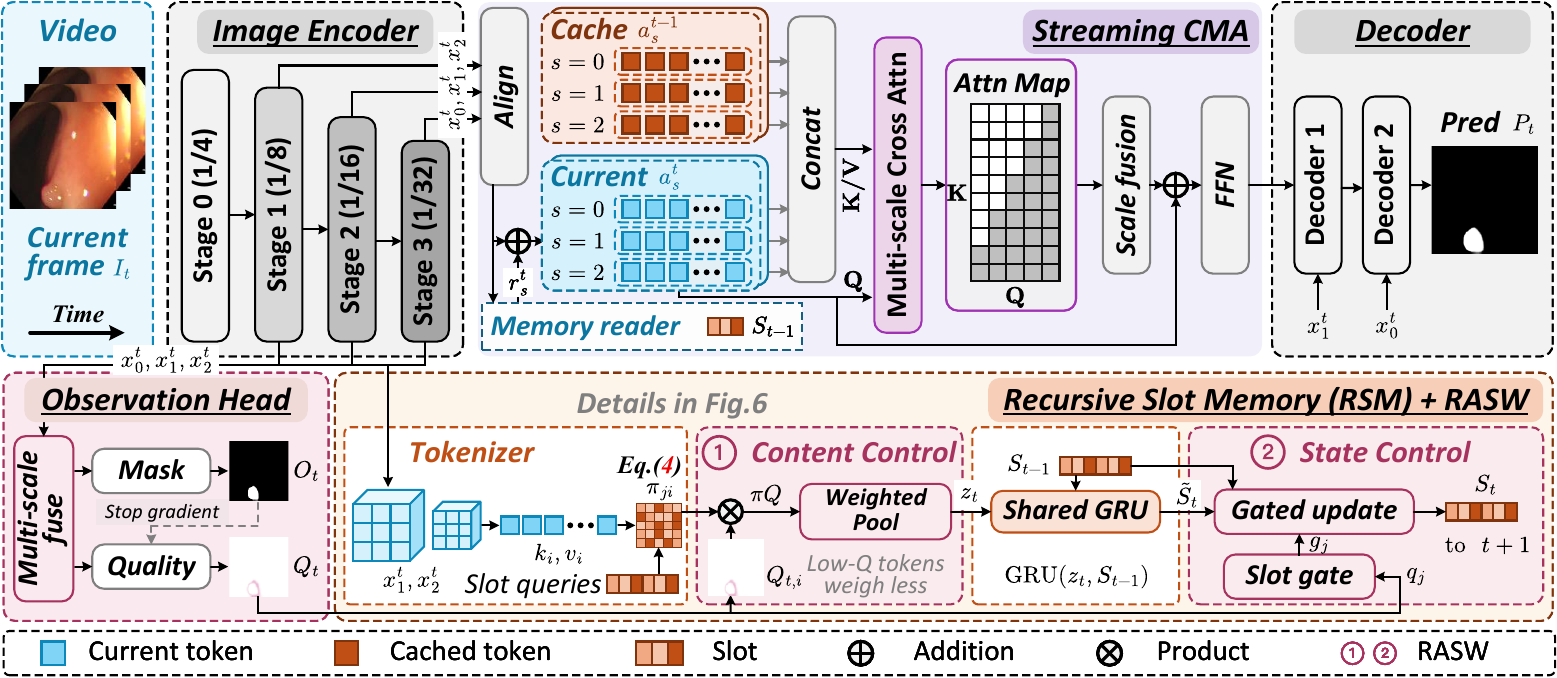}
\caption{\textbf{Overview of \ourmodel{}.} At frame $t$, the three-scale features of
$I_t$ read the old memory $S_{t-1}$, and the streaming CMA fuses the
memory-enhanced features with the cached previous-frame features before
decoding $P_t$. An observation head, which does not see the memory,
predicts the quality map $Q_t$. The memory tokenizer and a GRU of RSM
then compute a candidate state, and RASW uses $Q_t$ to control the
write at two points:
\textcircled{\scriptsize 1} the spatial pooling of candidate content and
\textcircled{\scriptsize 2} the per-slot acceptance gate, giving $S_t$
for the next frame. Gray: inherited from CMSA-Net; blue: streaming changes;
orange: RSM; red: RASW.}
\label{fig:framework}
\end{figure*}

\minisection{{\color{orange}\faLightbulb}~Observation 1: Long temporal
coverage need not require dense storage.} A short history can lose the
clear frames we need when degradation lasts long.
\textbf{\textit{Can history cover a long time without storing
every frame?}} We add blur, highlight, or occlusion intervals of
$D\in\{4,8,16,32\}$ frames. We compare dense FIFO banks
($K=2$--256) with sparse sampling. For $D=32$,
\cref{fig:obs1}(a) shows that FIFO8 drops from 0.5685 to 0.5206 when
its clean entries are pushed out. This is close to 0.5154 without
history. FIFO64 keeps 0.5586.
Storing more frames also costs more (\cref{tab:obs1}). Eight entries
sampled every four frames score 0.5541. This is close to dense FIFO32
(0.5547) with one quarter of the state (\cref{fig:obs1}(b)). A bank of
256 costs more and gives a lower Dice than FIFO64.
This motivates RSM (\cref{sec:slot}), which keeps a fixed number of
slots and updates them recursively.

\minisection{{\color{orange}\faLightbulb}~Observation 2: Partial degradation
calls for selective writes and retention.} \tight Long coverage also needs us to
keep useful history when we add new information.
\textbf{\textit{Should we write or skip a partly degraded frame
as a whole?}} We compare write rules with the same capacity ($K=8$
entries per location) in three settings: Follow (a region that follows the
polyp, area fraction 0.25), Scatter (six moving highlight spots, area
fraction 0.30), and Sweep (the target comes back from time to time).
All rules see the same degraded inputs; only the history writes differ.
When the target is occluded for a long time, skipping degraded frames raises
Dice from 0.2709 to 0.3270, and spatial selection with retention gives
the same gain (0.3266; \cref{tab:obs2}). With moving highlights,
skipping a whole frame also throws away usable regions: spatial selection
scores 0.4445, higher than skipping (0.4372) and unconditional writing
(0.4182). When the target comes back from time to time (Sweep), skipping
drops to nearly the no-history level (0.6099), while spatial selection
stays close to full writing. Across all 12 settings (three patterns $\times$
four lengths, $D\in\{16,32,64,128\}$), spatial selection is
within 0.5 points of the best rule, versus 5.8 for full writing and 2.6 for
skipping (\cref{fig:obs2}). These results use oracle masks. They show that
selective updates keep useful history. RASW (\cref{sec:quality}) learns
a spatial quality map to choose which content to write and how much each
slot updates.

\subsection{Overview of \ourmodel{}}
\label{sec:overview}
Guided by these observations, \ourmodel{} keeps CMA as its core, with
RSM providing its history and RASW protecting it (\cref{fig:framework}).
Given a video $\{I_t\}_{t=1}^{T}$, \ourmodel{} processes frames causally
and keeps RSM $S_t\in\mathbb{R}^{N\times d}$, whose $j$-th row $S_{t,j}$
is a $d$-dimensional slot and $S_0=\mathbf{0}$, and a cache of the
previous-frame encoder features. At time $t$, (1) the encoder extracts
three-scale features of $I_t$; (2) a memory reader queries $S_{t-1}$
with them; (3) the streaming CMA fuses the memory-enhanced features with
the cached ones, and the decoder predicts $P_t\in[0,1]^{H\times W}$;
(4) an observation head estimates an observation mask $O_t$ and a
quality map $Q_t$ from the current features alone; and (5) RASW writes
the frame into RSM under the control of $Q_t$, giving $S_t$. The
prediction reads only $S_{t-1}$, and the history is reset per video.
The encoder and decoder are inherited from CMSA-Net, and RSM has a fixed
size, so the cost per frame and the state do not grow with the video length.
The following subsections detail the streaming CMA, RSM, and RASW in turn.

\subsection{Streaming CMA}
\label{sec:cma}
\ourmodel{} keeps the alignment, the multi-scale attention, and the decoder
of CMA (\cref{sec:prelim}) and changes only its history: the
clip window and DMR references are replaced by the previous-frame cache
and the RSM readout.

\minisection{Fusion.} The memory reader (\cref{sec:slot}) returns a
readout $r_s^t\in\mathbb{R}^{C\times H_2\times W_2}$ for each scale. For each scale $s$, the keys and values
concatenate the cached previous-frame features and the memory-enhanced
current features as tokens,
$U_s^t=[a_s^{t-1};\,a_s^t+r_s^t]\in\mathbb{R}^{2H_2W_2\times C}$, with
maps flattened into one token per row. The current coarse feature
provides the query, and each scale provides the keys and values:
\begin{equation}
\mathbf{Q}=a_2^t W_Q,\qquad \mathbf{K}_s=U_s^t W_K,\qquad \mathbf{V}_s=U_s^t W_V,
\end{equation}
where $W_Q,W_K,W_V\in\mathbb{R}^{C\times C}$ are learnable projections ($\mathbf{Q}$ is the attention
query, not the quality map $Q_t$). The attention
results of the three scales are fused by weights $w=\mathrm{softmax}(\gamma)$
with learnable $\gamma\in\mathbb{R}^3$:
\begin{equation}
h_t=a_2^t+\sum_{s=0}^{2} w_s\,\mathrm{softmax}\Big(\frac{\mathbf{Q}\mathbf{K}_s^{\top}}{\sqrt{C}}\Big)\mathbf{V}_s,
\end{equation}
\begin{equation}
\bar h_t=h_t+\mathrm{FFN}(h_t),\qquad
\hat x_2^t=x_2^t+\bar h_t ,
\end{equation}
where FFN is a two-layer MLP. Layer normalization and multi-head splitting
are omitted for brevity, and the previous-frame term is dropped
at $t=1$. With $\bar h_t$ reshaped to $C\times H_2\times W_2$, the
CMSA-Net decoder turns $\hat x_2^t$ into $P_t$ and two deep supervision maps.

\minisection{Online processing.} Unlike CMA in CMSA-Net, the streaming
version queries only the current frame, takes its history from the cache
and RSM instead of the window and DMR, and caches the previous-frame
encoder features, so the encoder runs once per frame and each step reads
only the current frame, the cache, and $S_{t-1}$. Because the slots
enter CMA through $r_s^t$, the number of CMA tokens does not change
with the memory.

\subsection{Recursive Slot Memory (RSM)}
\label{sec:slot}
Unlike the frame-level history of Observation~1, whose cost grows with
the bank, RSM summarizes history in $N$ recursively updated slots, read
by CMA through $r_s^t$.

\minisection{Memory tokenizer.} The tokenizer turns the current frame into
$N$ candidate slot contents. The two coarser scales $x_1^t$ and $x_2^t$
are projected to $d$ channels, added with a scale embedding, and
flattened into $M=H_1W_1+H_2W_2$ spatial tokens with keys
$k_i\in\mathbb{R}^{1\times d}$, $i=1,\dots,M$. Their values
$v_i\in\mathbb{R}^{1\times d}$ are further conditioned on the soft
foreground/background probability $(p_i,1-p_i)$ of $P_t$ resized to
the token grid, so that slots can separate
polyp and background content without a hard mask. $N$ learnable slot
queries, shifted by a projection of the old state, attend to these
tokens:
\begin{equation}
\pi_{ji}=\mathrm{softmax}_i\!\left(\frac{(e_j+\bar S_{t-1,j}W_h)(k_iW_k)^{\top}}{\sqrt d}\right),
\end{equation}
where $e_j\in\mathbb{R}^{1\times d}$ is the $j$-th learnable query,
$j=1,\dots,N$, $\bar S_{t-1}=\mathrm{LN}(S_{t-1}W_m)$ is the projected
old state (LN: layer normalization), and $W_h,W_k,W_m\in\mathbb{R}^{d\times d}$; the shift is
omitted at $t=1$. The candidate content $z_t\in\mathbb{R}^{N\times d}$
pools the values with these weights (quality-weighted in \cref{sec:quality}).

\minisection{Recursive update.} A shared GRU cell~\citep{cho2014gru} combines the
candidate content with the old state slot by slot,
$\tilde S_t=\mathrm{GRU}(z_t,S_{t-1})\in\mathbb{R}^{N\times d}$.
The slot gate of RASW (\cref{sec:quality}) then decides how
much of $\tilde S_t$ is accepted.

\minisection{Memory reader.} The three-scale current features, resized to
$H_2\times W_2$ and projected to $d$ channels, query the slots
$\bar S_{t-1}$ with multi-head cross-attention shared across scales,
and a per-scale linear layer maps the readout back to $C$ channels,
giving $r_s^t$ for CMA, with $r_s^t=\mathbf{0}$ at $t=1$.

\minisection{Budget.} The memory holds $Nd$ values regardless of the
video length, and the previous-frame cache holds
$C(H_0W_0+H_1W_1+H_2W_2)$ values, so the full state is constant and
reading costs $O(H_2W_2N)$ per scale (\cref{tab:efficiency}).

\begin{figure}[pos=!t]
\centering
\includegraphics[width=\columnwidth]{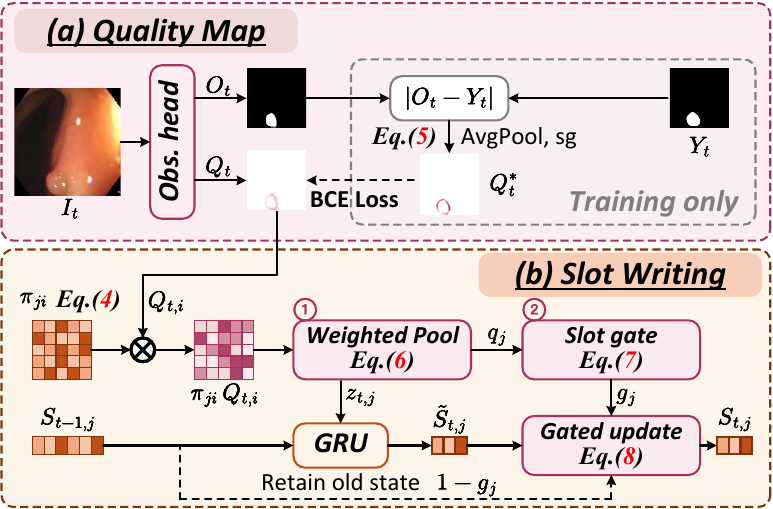}
\caption{\textbf{Reliability-aware slot writing (RASW).} (a) The quality map is supervised by
the local error of the observation branch. (b) In one write, the quality
map controls \textcircled{\scriptsize 1} which locations form the
candidate content and \textcircled{\scriptsize 2} how much of the GRU
update each slot accepts.}
\label{fig:write}
\end{figure}

\subsection{Reliability-Aware Slot Writing (RASW)}
\label{sec:quality}
Because CMA reads RSM at every later frame and a degraded write harms
memory during degradation (Observation~2), RASW estimates the reliability of
the current observation as a pixel-level quality map that controls
each write at two points (\cref{fig:write}).

\minisection{Observation head.} It fuses the three-scale
features $x_0^t,x_1^t,x_2^t$ \emph{before} any memory readout and
predicts an observation mask $O_t=\sigma(o_t)$ and a spatial quality
map $Q_t=\sigma(\ell_t)$, where $o_t,\ell_t\in\mathbb{R}^{H_0\times W_0}$
are logits, $\sigma$ is the sigmoid function, and the quality branch
also takes the detached $O_t$, so $Q_t$ describes the current
observation alone and cannot inherit errors stored in the memory.

\minisection{Quality target.} \tight $Q_t$ is supervised by the complement of the
local prediction error of the observation branch:
\begin{equation}
Q_t^{*}=\mathrm{sg}\!\left[1-\mathrm{AvgPool}_{\kappa\times\kappa}\big(|O_t-Y_t|\big)\right],
\end{equation}
where $\kappa$ is the window size, $O_t$ is upsampled to $H\times W$, $Y_t\in\{0,1\}^{H\times W}$
is the ground-truth mask, and $\mathrm{sg}$ denotes stop gradient, so the observation branch is not pushed toward the quality
target. $Q_t$ therefore estimates how reliable the current observation
is at each location, i.e., the reliability of its write.

\minisection{Content control.} The first control point decides which
locations form the candidate content. The tokenizer weights the
attention by the quality of each token and renormalizes:
\begin{equation}
z_{t,j}=\mathrm{LN}\!\left(\frac{\sum_i \pi_{ji}\,Q_{t,i}\,v_iW_v}{\max\big(\sum_i \pi_{ji}\,Q_{t,i},\,\epsilon\big)}\right),
\end{equation}
where $Q_{t,i}$ is $Q_t$ resized to the token grid at token $i$,
$W_v\in\mathbb{R}^{d\times d}$, and $\epsilon$ is a small constant.
Low-quality locations thus contribute less, and the normalization lets
this term decide \emph{where} content comes from, not how much is
accepted.

\minisection{State control.} The second control point decides how much of
the update each slot accepts. Pooling $Q_t$ with the original attention
gives the quality of slot $j$, $q_j=\sum_i \pi_{ji}Q_{t,i}\in[0,1]$, from which we
compute its write gate and update the state:
\begin{gather}
g_j=\mathrm{clip}(q_j,0,1)\cdot\sigma(\alpha q_j+\beta), \\
S_{t,j}=S_{t-1,j}+g_j\big(\tilde S_{t,j}-S_{t-1,j}\big),
\end{gather}
where $\alpha$ and $\beta$ are learned scalars. Unlike a frame-level
gate, $g_j$ lets slots on low-quality locations retain more old state.

A local highlight mainly calls for the content control,
which takes the content from the clear part of the frame, whereas
whole-frame blur leaves no clear part and calls for the state control,
which keeps the old state. $Q_t$ is detached during writing and learns from
its own supervision.

\subsection{Training Objective}
\label{sec:loss}
The total loss is
\begin{equation}
\mathcal{L}=\mathcal{L}_{\mathrm{seg}}+\lambda_o\mathcal{L}_{\mathrm{obs}}+\lambda_q\mathcal{L}_{\mathrm{qual}},
\end{equation}
where $\lambda_o$ and $\lambda_q$ are loss weights. $\mathcal{L}_{\mathrm{seg}}$
sums the weighted binary cross-entropy, weighted IoU, and Dice losses on
the final prediction and the two deep supervision outputs.
$\mathcal{L}_{\mathrm{obs}}$ applies the same loss to $O_t$.
$\mathcal{L}_{\mathrm{qual}}$ is the binary cross-entropy between $Q_t$
and $Q_t^{*}$, averaged separately over the polyp interior, a boundary
band ($\kappa\times\kappa$ dilation minus erosion), and the background, and then
averaged over the non-empty regions, so that the large background does
not dominate.

\minisection{Streaming training.} \tight The model is trained on long sequences
processed frame by frame with truncated backpropagation through time:
gradients are truncated at fixed intervals, while the memory and the
cache are carried over. To expose the memory to degraded inputs, a
sequence may receive a continuous interval of corrupted frames, either
Gaussian blur or a synthetic specular highlight at a fixed position.

\section{Experiments}
\label{sec:experiments}

\begin{table*}[pos=!t]
\centering
\caption{Quantitative comparison on the SUN-SEG-Easy dataset. All methods are retrained under the same settings. The best and second-best results are highlighted in \textbf{bold} and \underline{underlined}, respectively.}
\label{tab:easy}
\tablesetup
\begin{tabular*}{\linewidth}{@{\extracolsep{\fill}}lllcccccccccccc@{}}
\toprule
\multirow{2}{*}{Method} & \multirow{2}{*}{Publication} & \multirow{2}{*}{Backbone} & \multicolumn{6}{c}{SUN-SEG-Easy-Seen (\%)} & \multicolumn{6}{c}{SUN-SEG-Easy-Unseen (\%)} \\
\cmidrule(lr){4-9}\cmidrule(lr){10-15}
 & & & $S_\alpha\uparrow$ & $E_\phi^{mn}\uparrow$ & $F_\beta^w\uparrow$ & Dice$\uparrow$ & IoU$\uparrow$ & MAE$\downarrow$ & $S_\alpha\uparrow$ & $E_\phi^{mn}\uparrow$ & $F_\beta^w\uparrow$ & Dice$\uparrow$ & IoU$\uparrow$ & MAE$\downarrow$ \\
\midrule
PraNet & MICCAI'20 & Res2Net-50 & 91.8 & 94.2 & 87.7 & 88.3 & 82.5 & 2.0 & 78.1 & 78.8 & 66.3 & 66.5 & 58.2 & 5.2 \\
ACSNet & MICCAI'20 & ResNet-34 & 92.0 & 94.2 & 87.4 & 88.2 & 82.8 & 1.7 & 77.2 & 76.6 & 63.0 & 63.8 & 56.4 & 4.6 \\
SANet & MICCAI'21 & Res2Net-50 & 91.6 & 93.3 & 86.6 & 87.2 & 82.0 & 1.8 & 75.0 & 72.8 & 59.0 & 59.3 & 52.4 & 5.2 \\
SEPNet & TCSVT'24 & PVTv2-B2 & 93.1 & 96.2 & 88.3 & 89.6 & 83.4 & 1.7 & 82.9 & 88.3 & 73.5 & 75.1 & 66.6 & 4.2 \\
\midrule
PNS-Net & MICCAI'21 & Res2Net-50 & 90.6 & 91.0 & 83.6 & 84.1 & 78.3 & 2.0 & 76.7 & 74.4 & 61.6 & 61.8 & 54.5 & 4.8 \\
PNS+ & MIR'22 & Res2Net-50 & 91.7 & 92.5 & 84.8 & 85.5 & 78.7 & 2.1 & 80.6 & 79.8 & 67.6 & 67.8 & 59.1 & 4.4 \\
MAST & AIIM'25 & PVTv2-B2 & 92.5 & 96.2 & 87.8 & 89.3 & 82.7 & 1.6 & 83.2 & 89.4 & 74.9 & 77.0 & 67.4 & 3.7 \\
SALI & MICCAI'24 & PVTv2-B2 & 90.2 & 93.2 & 84.9 & 85.8 & 78.9 & 2.4 & 73.1 & 75.2 & 58.7 & 59.2 & 50.2 & 6.3 \\
SALI & MICCAI'24 & PVTv2-B5 & 90.7 & 93.7 & 85.1 & 86.2 & 79.6 & 2.2 & 77.1 & 82.1 & 64.6 & 65.6 & 56.8 & 5.5 \\
STDDNet & ICCV'25 & Res2Net-50 & 93.5 & 96.0 & 89.7 & 90.5 & 85.0 & 1.5 & 81.7 & 83.0 & 72.1 & 72.4 & 64.3 & 3.7 \\
STDDNet & ICCV'25 & PVTv2-B2 & 94.1 & 96.9 & 90.5 & 91.5 & 86.1 & 1.4 & 86.0 & 90.3 & 78.6 & 80.1 & 72.4 & 3.4 \\
CMSA-Net & MICCAI'26 & Res2Net-50 & 94.5 & 97.3 & 90.5 & 91.9 & 86.5 & 1.2 & 84.4 & 90.1 & 75.3 & 77.5 & 69.2 & 3.5 \\
CMSA-Net & MICCAI'26 & PVTv2-B2 & \textbf{95.1} & \underline{97.5} & \underline{91.6} & \underline{92.6} & \underline{87.6} & \underline{1.1} & 86.7 & 90.3 & 79.3 & 80.3 & 72.6 & 2.9 \\
\midrule
\rowcolor{c1} \ourmodel{} (Ours) & -- & Res2Net-50 & \underline{94.6} & 97.4 & 91.1 & 92.4 & 87.1 & 1.2 & \underline{88.6} & \underline{92.4} & \underline{81.8} & \underline{83.8} & \underline{76.8} & \underline{2.5} \\
\rowcolor{c1} \ourmodel{} (Ours) & -- & PVTv2-B2 & \textbf{95.1} & \textbf{97.6} & \textbf{91.9} & \textbf{92.9} & \textbf{88.0} & \textbf{1.0} & \textbf{89.4} & \textbf{93.7} & \textbf{83.6} & \textbf{85.0} & \textbf{78.1} & \textbf{2.2} \\
\bottomrule
\end{tabular*}
\end{table*}
\begin{table*}[pos=!t]
\centering
\caption{Quantitative comparison on the SUN-SEG-Hard dataset. All methods are retrained under the same settings. The best and second-best results are highlighted in \textbf{bold} and \underline{underlined}, respectively.}
\label{tab:hard}
\tablesetup
\begin{tabular*}{\linewidth}{@{\extracolsep{\fill}}lllcccccccccccc@{}}
\toprule
\multirow{2}{*}{Method} & \multirow{2}{*}{Publication} & \multirow{2}{*}{Backbone} & \multicolumn{6}{c}{SUN-SEG-Hard-Seen (\%)} & \multicolumn{6}{c}{SUN-SEG-Hard-Unseen (\%)} \\
\cmidrule(lr){4-9}\cmidrule(lr){10-15}
 & & & $S_\alpha\uparrow$ & $E_\phi^{mn}\uparrow$ & $F_\beta^w\uparrow$ & Dice$\uparrow$ & IoU$\uparrow$ & MAE$\downarrow$ & $S_\alpha\uparrow$ & $E_\phi^{mn}\uparrow$ & $F_\beta^w\uparrow$ & Dice$\uparrow$ & IoU$\uparrow$ & MAE$\downarrow$ \\
\midrule
PraNet & MICCAI'20 & Res2Net-50 & 88.4 & 91.9 & 83.1 & 83.9 & 76.6 & 3.1 & 78.7 & 80.2 & 66.7 & 67.5 & 58.7 & 5.3 \\
ACSNet & MICCAI'20 & ResNet-34 & 87.2 & 91.0 & 80.6 & 82.0 & 74.8 & 3.6 & 76.2 & 77.6 & 61.0 & 62.4 & 54.7 & 5.3 \\
SANet & MICCAI'21 & Res2Net-50 & 87.4 & 90.5 & 81.0 & 82.0 & 74.8 & 3.3 & 75.3 & 73.6 & 59.0 & 59.5 & 52.7 & 5.5 \\
SEPNet & TCSVT'24 & PVTv2-B2 & 89.4 & 94.0 & 83.5 & 85.7 & 77.6 & 3.4 & 84.7 & 89.5 & 74.5 & 77.4 & 68.4 & 3.9 \\
\midrule
PNS-Net & MICCAI'21 & Res2Net-50 & 87.0 & 89.2 & 78.7 & 79.6 & 72.1 & 3.3 & 76.7 & 75.5 & 60.9 & 61.5 & 53.9 & 5.0 \\
PNS+ & MIR'22 & Res2Net-50 & 88.7 & 90.2 & 80.6 & 81.3 & 72.8 & 3.0 & 79.8 & 79.3 & 65.4 & 66.1 & 57.1 & 5.0 \\
MAST & AIIM'25 & PVTv2-B2 & 89.2 & 94.2 & 83.2 & 85.3 & 76.7 & 2.6 & 85.6 & 91.3 & 77.2 & 79.9 & 70.8 & 3.1 \\
SALI & MICCAI'24 & PVTv2-B2 & 86.8 & 90.9 & 79.9 & 81.0 & 72.6 & 3.4 & 72.8 & 75.9 & 56.9 & 57.9 & 48.7 & 6.8 \\
SALI & MICCAI'24 & PVTv2-B5 & 86.6 & 91.0 & 79.7 & 81.0 & 72.9 & 3.8 & 76.5 & 81.3 & 62.0 & 63.6 & 54.7 & 5.7 \\
STDDNet & ICCV'25 & Res2Net-50 & 91.3 & 95.2 & 86.9 & 88.1 & 81.0 & 2.3 & 83.4 & 85.6 & 74.1 & 75.0 & 67.3 & 3.7 \\
STDDNet & ICCV'25 & PVTv2-B2 & 91.1 & 95.0 & 86.0 & 87.8 & 80.6 & 2.8 & 86.3 & 90.2 & 78.1 & 80.2 & 72.2 & 3.5 \\
CMSA-Net & MICCAI'26 & Res2Net-50 & 92.7 & 96.1 & 87.1 & 89.8 & 83.0 & 1.8 & 85.1 & 89.8 & 75.0 & 78.0 & 69.5 & 3.6 \\
CMSA-Net & MICCAI'26 & PVTv2-B2 & 92.3 & 95.6 & 87.1 & 88.9 & 82.1 & 1.9 & 87.3 & 91.0 & 79.6 & 81.3 & 73.7 & 2.9 \\
\midrule
\rowcolor{c1} \ourmodel{} (Ours) & -- & Res2Net-50 & \underline{92.9} & \underline{96.6} & \underline{88.3} & \underline{90.2} & \underline{83.4} & \underline{1.6} & \underline{88.7} & \underline{93.3} & \underline{81.5} & \underline{84.0} & \underline{76.5} & \underline{2.4} \\
\rowcolor{c1} \ourmodel{} (Ours) & -- & PVTv2-B2 & \textbf{93.3} & \textbf{96.7} & \textbf{89.0} & \textbf{90.6} & \textbf{84.2} & \textbf{1.3} & \textbf{89.9} & \textbf{94.0} & \textbf{83.5} & \textbf{85.7} & \textbf{78.5} & \textbf{2.1} \\
\bottomrule
\end{tabular*}
\end{table*}

\begin{figure*}[pos=!t]
\centering
\includegraphics[width=\textwidth]{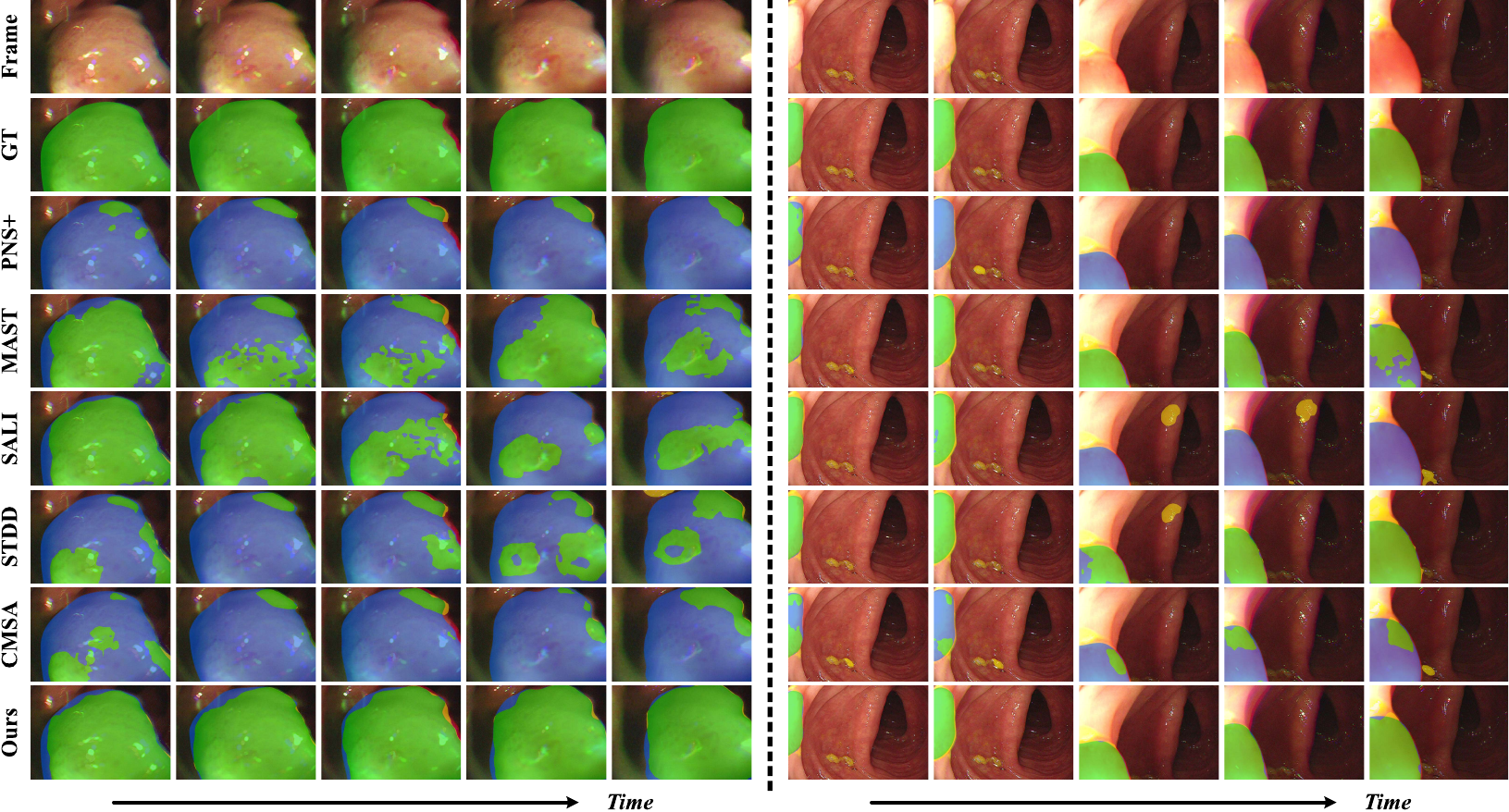}
\caption{\textbf{Qualitative comparison on SUN-SEG-Hard-Unseen.} Left (Case~1,
from \texttt{case32\_4}): consecutive frames with blur and specular
highlights. Right (Case~2, from \texttt{case42}): a polyp with a large scale
change. Green, yellow, and blue
denote true positives, false positives, and false negatives.}
\label{fig:qualitative}
\end{figure*}

\subsection{Experimental Settings}
\label{sec:setup}
\minisection{Datasets.} We evaluate on SUN-SEG~\citep{sunseg}, the largest
VPS benchmark. Following prior work~\citep{sunseg,chen2025stddnet,cmsa}, the
training set contains 112 clips with 19,544 frames, and the test set is
divided into four subsets: Easy-Seen (33 clips, 4,719 frames),
Easy-Unseen (86 clips, 12,351 frames), Hard-Seen (17 clips, 3,882
frames), and Hard-Unseen (37 clips, 8,640 frames). Easy and Hard denote
segmentation difficulty, and Seen and Unseen indicate whether the test
videos overlap with the training videos. We further evaluate on
CVC-ClinicDB (CVC-612)~\citep{bernal2015cvc}, which contains 612 frames
at a resolution of $384\times288$. We use all 612 images as an unseen
test set for SUN-SEG-trained models, without fine-tuning on
CVC-ClinicDB.

\minisection{Evaluation metrics.} Following CMSA-Net~\citep{cmsa}, we report six metrics:
structure-measure $S_\alpha$~\citep{fan2017smeasure}, enhanced-alignment
measure $E_\phi^{mn}$~\citep{fan2021emeasure}, weighted F-measure
$F_\beta^w$~\citep{margolin2014fbw}, Dice, IoU, and mean absolute error
(MAE), computed with the evaluation code of CMSA-Net. Larger values are
better for all metrics except MAE.
For efficiency, we report GFLOPs, number of parameters, inference speed (FPS), and cross-frame state size.
Because frames within a video are not independent, we test statistical significance per
video: each video is scored by its mean frame Dice, and \ourmodel{} is
compared with CMSA-Net, STDDNet, SEPNet, PNS+, and PNS-Net on the same videos using a
two-sided Wilcoxon signed-rank test, reporting both the mean difference and its
95\% confidence interval (CI).

\minisection{Compared methods.} We compare \ourmodel{} with 10 representative state-of-the-art methods
(\cref{tab:easy,tab:hard}),
all retrained on the SUN-SEG training set under the same protocol:
(1) image-based polyp segmentation methods, including
PraNet~\citep{fan2020pranet}, ACSNet~\citep{zhang2020acsnet},
SANet~\citep{wei2021sanet}, and SEPNet~\citep{wang2024sepnet}; and (2)
video polyp segmentation methods, including PNS-Net~\citep{ji2021pnsnet},
PNS+~\citep{sunseg}, MAST~\citep{chen2025mast}, SALI~\citep{hu2024sali},
STDDNet~\citep{chen2025stddnet}, and CMSA-Net~\citep{cmsa}.

\minisection{Implementation details.} Following CMSA-Net~\citep{cmsa}, we implement \ourmodel{} with two ImageNet-pretrained
encoders, Res2Net-50~\citep{gao2021res2net} and PVTv2-B2~\citep{wang2022pvtv2},
using the same configuration for both: $C=32$
channels, $4$ attention heads ($d_h=C/4$), an FFN width of $4C$, and
$N=32$ slots of dimension $d=64$. Inputs are resized to $352\times352$. We set $\kappa=7$ and $\epsilon=10^{-6}$, initialize the gate
parameters to $\alpha=2$ and $\beta=-1$, and use $\lambda_o=0.5$ and
$\lambda_q=0.2$. Training sequences have $36$ frames with gradients
truncated every $6$ frames, and with probability $0.3$ a sequence
receives $2$ to $12$ consecutive corrupted frames. We use AdamW with a learning rate
of $2\times10^{-4}$ ($0.1\times$ for the encoder), a weight decay of
$10^{-4}$, a batch size of $8$, and cosine annealing~\citep{loshchilov2016sgdr} with a period of $20$ epochs
and a minimum of $5\times10^{-5}$, for 30 epochs, following
STDDNet~\citep{chen2025stddnet} and CMSA-Net~\citep{cmsa}. The batch normalization layers
of the encoder are frozen, and the new modules use group normalization. All \ourmodel{}
variants are trained on the official SUN-SEG training set; the 70:30
case split is used only for the observations in
\cref{sec:observation}. On SUN-SEG, inference processes each video
in temporal order from its first frame without future frames.
All experiments are run on a single NVIDIA A100 GPU with 40~GB of memory.

\subsection{Comparison with State-of-the-art Methods}
\label{sec:sota}
\minisection{Quantitative comparison.} \cref{tab:easy,tab:hard} report the results on SUN-SEG-Easy and
SUN-SEG-Hard, where \ourmodel{} ranks first on all four test sets. With
PVTv2-B2, \ourmodel{} improves the Dice of CMSA-Net from 80.3\% to 85.0\%
on Easy-Unseen and from 81.3\% to 85.7\% on Hard-Unseen. The gains on
unseen videos indicate that RSM and RASW keep reliable context through
long intervals of blur, highlights, and low contrast. With Res2Net-50,
the average Dice rises from 84.3\% to 87.6\%, so the gain is
encoder-agnostic. CMA thus needs long, bounded, reliable history.

\minisection{Statistical significance.} Over all 173 test videos (PVTv2-B2),
\ourmodel{} improves the per-video Dice of CMSA-Net by 3.5\% (95\% CI
[2.6, 4.4], $p<10^{-17}$) and of STDDNet by 4.2\% (95\% CI [2.9, 5.4],
$p<10^{-13}$). On both unseen sets the gains over all five tested methods
are significant ($p<10^{-5}$). On the seen sets, where CMSA-Net and
STDDNet already exceed 87\% Dice, the gains are smaller and not always
significant (e.g., $p=0.07$ against CMSA-Net on Easy-Seen and $p=0.19$
against STDDNet on Hard-Seen), so the gains come mainly from unseen
videos (\cref{sec:significance}).

\minisection{Qualitative comparison.} \cref{fig:qualitative} shows two
difficult cases. In Case~1, the compared methods miss most of the polyp
on consecutive blurred or highlighted frames, while \ourmodel{} keeps an
accurate mask because RSM holds the polyp content written before the
interval. In Case~2, the polyp grows as the camera approaches;
\ourmodel{} follows the scale change, whereas the others under-segment
the enlarged polyp.

\begin{table*}[pos=!t]
\centering
\caption{Dice (\%) on scenario subsets of the SUN-SEG test sets. Left:
subsets that call for long-range history; right: subsets with spatially
non-uniform degradation. SV, IB, FM, GH, and OCC are official SUN-SEG
video attributes; the other subsets are defined in
\cref{sec:scenario}. The last three rows add the components of \ourmodel{} one at a time
(\cref{sec:ablation}). All methods use PVTv2-B2 except PNS+ (Res2Net-50); $\Delta$: \ourmodel{} $-$ CMSA-Net.}
\label{tab:scenario}
\tablesetup
\begin{tabular*}{\linewidth}{@{\extracolsep{\fill}}lcccccccccccc@{}}
\toprule
\multirow{2}{*}{Method} & \multirow{2}{*}{All} & \multicolumn{5}{c}{Long-range history} & \multicolumn{6}{c}{Non-uniform degradation} \\
\cmidrule(lr){3-7}\cmidrule(lr){8-13}
 & & SV & IB & ScaleChg & LowCon & Late & FM & GH & OCC & Blur & Highl. & PostDeg \\
\midrule
\#Frames & 29{,}592 & 6{,}632 & 5{,}988 & 2{,}787 & 2{,}960 & 5{,}074 & 5{,}529 & 13{,}062 & 8{,}729 & 2{,}960 & 2{,}960 & 1{,}995 \\
\midrule
SEPNet & 79.7 & 72.4 & 63.1 & 72.6 & 67.6 & 81.8 & 71.8 & 80.7 & 78.9 & 72.8 & 77.6 & 75.2 \\
PNS+ & 72.1 & 61.8 & 48.0 & 63.1 & 55.2 & 73.6 & 63.9 & 73.2 & 71.6 & 63.4 & 70.9 & 66.4 \\
STDDNet & 83.2 & 76.7 & 68.8 & 75.6 & 71.2 & 84.7 & 75.0 & 83.5 & 82.0 & 75.4 & 81.3 & 78.8 \\
CMSA-Net & 84.0 & 75.9 & 68.5 & 76.2 & 72.9 & 86.6 & 76.2 & 84.2 & 83.9 & 77.6 & 83.2 & 81.1 \\
\midrule
\circlednum{1} Streaming CMA, no memory & 78.9 & 70.7 & 62.9 & 70.7 & 68.4 & 82.0 & 70.8 & 79.0 & 77.1 & 71.9 & 76.5 & 74.8 \\
\circlednum{2} + RSM & 84.6 & 77.6 & 70.4 & 77.9 & 74.3 & 86.0 & 76.5 & 84.8 & 85.2 & 80.7 & 83.6 & 82.7 \\
\circlednum{3} + RASW (\ourmodel{}) & \textbf{87.4} & \textbf{81.1} & \textbf{75.5} & \textbf{80.7} & \textbf{76.9} & \textbf{88.7} & \textbf{79.2} & \textbf{87.3} & \textbf{87.6} & \textbf{83.7} & \textbf{86.5} & \textbf{85.5} \\
\midrule
$\Delta$                           & +3.4 & +5.2 & +7.0 & +4.5 & +4.0 & +2.1 & +3.0 & +3.1 & +3.7 & +6.1 & +3.3 & +4.4 \\
\bottomrule
\end{tabular*}
\end{table*}

\subsection{Scenario-Based Analysis}
\label{sec:scenario}
To examine where the gains come from, we evaluate on subsets of the four
test sets tied to the two challenges in \cref{sec:introduction}
(\cref{tab:scenario}). Subsets
are defined from images and ground-truth masks only, with thresholds at
the 10\% quantile over all test frames, independent of any
prediction; \cref{sec:observation} applies the same rules to the
held-out 30\% split of the SUN-SEG training set.

\minisection{Subsets.} \tight For \emph{long-range history}, we use the
official SUN-SEG video attributes scale variation (SV) and indefinable
boundaries (IB)~\citep{sunseg}, and three frame-level subsets: ScaleChg,
the 10\% of frames with the largest $|\log(A_t/A_{t-10})|$, where $A_t$
is the annotated polyp area; LowCon, the 10\% of frames with the
smallest CIE-Lab distance between the polyp and a surrounding ring; and
Late, frames after the first 150 of a video. For \emph{spatially
non-uniform degradation}, we use the official attributes fast motion
(FM), ghosting (GH), and occlusion (OCC), and three frame-level
subsets: Blur, the 10\% of frames with the lowest Laplacian variance;
Highlight (Highl.), the 10\% of frames with the largest fraction of specular pixels;
and PostDeg, the first five clear frames after at least two consecutive
Blur or Highlight frames. PostDeg measures how much a degraded interval
affects the frames that follow it.

\minisection{Results.} \textbf{1) \ourmodel{} improves most where the
challenges arise.} Over CMSA-Net, \ourmodel{} improves Dice from 84.0\%
to 87.4\% on all frames, with average gains of 4.6\% on the long-range
subsets and 3.9\% on the degradation subsets, and the largest gains on
IB (68.5\% to 75.5\%) and Blur (77.6\% to 83.7\%). \textbf{2) RSM and RASW are
complementary.} RSM raises the average Dice of the streaming model by
6.3\% on the long-range subsets and 7.2\% on the degradation subsets,
most on Blur (71.9\% to 80.7\%) and OCC (77.1\% to 85.2\%): when the current frame is
unreliable, the slots still hold earlier clear content. RASW adds a
further 3.3\% and 2.7\%, improving every subset by at least 2.4\%. On
Late, RASW lifts the Dice from 86.0\% (RSM alone) to 88.7\%, above
CMSA-Net (86.6\%), suggesting that unchecked low-quality updates
accumulate over long videos.

\subsection{Generalization and Efficiency}
\label{sec:general}
\begin{table}[pos=!t]
\centering
\caption{Cross-dataset comparison on CVC-ClinicDB (CVC-612). The best and second-best results are highlighted in \textbf{bold} and \underline{underlined}, respectively. R: Res2Net-50; P: PVTv2-B2.}
\label{tab:cvc}
\tablesetup
\begin{tabular*}{\linewidth}{@{\extracolsep{\fill}}lcccccc@{}}
\toprule
Method & $S_\alpha\uparrow$ & $E_\phi^{mn}\uparrow$ & $F_\beta^w\uparrow$ & Dice$\uparrow$ & IoU$\uparrow$ & MAE$\downarrow$ \\
\midrule
PraNet      & 84.3 & 86.7 & 76.0 & 76.4 & 68.0 & 3.9 \\
SEPNet     & 85.2 & 88.1 & 77.6 & 77.4 & 69.9 & 3.8 \\
STDDNet (R) & 87.0 & 88.8 & 80.3 & 80.4 & 72.5 & 3.2 \\
STDDNet (P) & 88.7 & \underline{91.8} & 83.7 & 83.7 & 75.7 & \underline{2.8} \\
CMSA-Net (R)           & 82.6 & 83.4 & 74.1 & 73.4 & 65.1 & 4.4 \\
CMSA-Net (P)           & 87.2 & 89.0 & 80.1 & 80.4 & 72.1 & 3.4 \\
\midrule
\ourmodel{} (R)                    & \textbf{89.7} & \textbf{93.5} & \textbf{85.1} & \textbf{86.1} & \textbf{78.5} & \textbf{2.6} \\
\ourmodel{} (P)                    & \underline{89.0} & \underline{91.8} & \underline{83.8} & \underline{84.3} & \underline{76.4} & \underline{2.8} \\
\bottomrule
\end{tabular*}
\end{table}

\begin{table}[pos=!t]
\centering
\caption{Efficiency comparison over 6 frames (one clip for clip-based methods).
State: memory carried between frames at $352\times352$ (N/A: image-based).
R: Res2Net-50; P: PVTv2-B2.}
\label{tab:efficiency}
\tablesetup
\begin{tabular*}{\linewidth}{@{\extracolsep{\fill}}lcccc@{}}
\toprule
Method & GFLOPs$\downarrow$ & Param. (M)$\downarrow$ & FPS$\uparrow$ & State (MB)$\downarrow$ \\
\midrule
PraNet       & 78.89  & 30.50 & 42 & N/A \\
ACSNet     & 130.52 & 29.45 & 21 & N/A \\
SALI            & 86.58  & 82.73 & 10 & 7.46 \\
STDDNet (R)& 75.17  & 38.16 & 45 & 7.09 \\
STDDNet (P)& 64.18  & 29.67 & 36 & 7.09 \\
CMSA-Net (R)          & 73.36  & 29.61 & 47 & 7.09 \\
CMSA-Net (P)          & 59.42  & 25.79 & 38 & 7.09 \\
\midrule
\ourmodel{} (R)                        & 73.73 & 29.68 & 50 & 0.32 \\
\ourmodel{} (P)                        & 59.84 & 25.88 & 48 & 0.32 \\
\bottomrule
\end{tabular*}
\end{table}

\minisection{Results on CVC-ClinicDB.} \Cref{tab:cvc} reports
cross-dataset results with the backbones of \cref{tab:easy}. With
PVTv2-B2, \ourmodel{} reaches a Dice of 84.3\%, 3.9\% higher than
CMSA-Net on this unseen dataset. Tested over the 29 sequences of
CVC-ClinicDB, the gains of \ourmodel{} (PVTv2-B2) are significant over
CMSA-Net (+4.1\%, 95\% CI [1.1, 7.1], $p<10^{-3}$) and over SEPNet
(+7.6\%, 95\% CI [1.9, 13.4], $p<10^{-4}$).

\minisection{Efficiency.} \cref{tab:efficiency} compares the
computational cost. Running once per frame and reusing cached features
instead of re-encoding a clip, \ourmodel{} reaches 48 FPS with 25.88~M parameters and 9.97 GFLOPs per
frame (PVTv2-B2). Its state between frames is 0.32~MB and stays constant over the
video, whereas SALI needs 7.46~MB to keep 35 past frames.

\subsection{Ablation Study}
\label{sec:ablation}
All ablations use PVTv2-B2 with the same training set, schedule, and
losses, change only the indicated memory or write mechanism, and report
Dice on the test sets.

\minisection{Step-wise ablation.} \tight \cref{tab:abl_step} adds the
components cumulatively. \circlednum{1} turns CMSA-Net into a streaming
model with only the previous-frame cache, lowering the average Dice over
the four test sets from 85.8\% to 81.3\% with single-frame input.
\circlednum{3} adds RSM and recovers 85.8\%, with the largest gains on
the unseen sets (e.g., Easy-Unseen from 74.1\% to 81.8\%), because the
slots carry context beyond the previous frame. \circlednum{5} adds RASW
and reaches 88.6\%, improving all four test sets.

\begin{table}[pos=!t]
\centering
\caption{Step-wise ablation and RSM slot capacity (Dice, \%).
Slot-count variants disable RASW; the full model uses 32 slots.}
\label{tab:abl_step}
\tablesetup
\begin{tabular*}{\linewidth}{@{\extracolsep{\fill}}lccccc@{}}
\toprule
\multirow{2}{*}{Model} & State & \multirow{2}{*}{Easy-S} & \multirow{2}{*}{Easy-U} & \multirow{2}{*}{Hard-S} & \multirow{2}{*}{Hard-U} \\
 & (MB) & & & & \\
\midrule
\circlednum{1} Streaming only         & 0.310 & 90.2 & 74.1 & 85.1 & 75.7 \\
\midrule
\circlednum{2} + RSM, $N{=}16$                     & 0.314 & 91.6 & 76.3 & 87.0 & 78.3 \\
\circlednum{3} + RSM, $N{=}32$        & 0.318 & 91.2 & 81.8 & 87.9 & 82.2 \\
\circlednum{4} + RSM, $N{=}64$                     & 0.326 & 91.6 & 82.1 & 87.8 & 81.1 \\
\midrule
\circlednum{5} + RASW (\ourmodel{})   & 0.318 & \textbf{92.9} & \textbf{85.0} & \textbf{90.6} & \textbf{85.7} \\
\bottomrule
\end{tabular*}
\end{table}

\begin{figure}[pos=!t]
\centering
\includegraphics[width=\columnwidth]{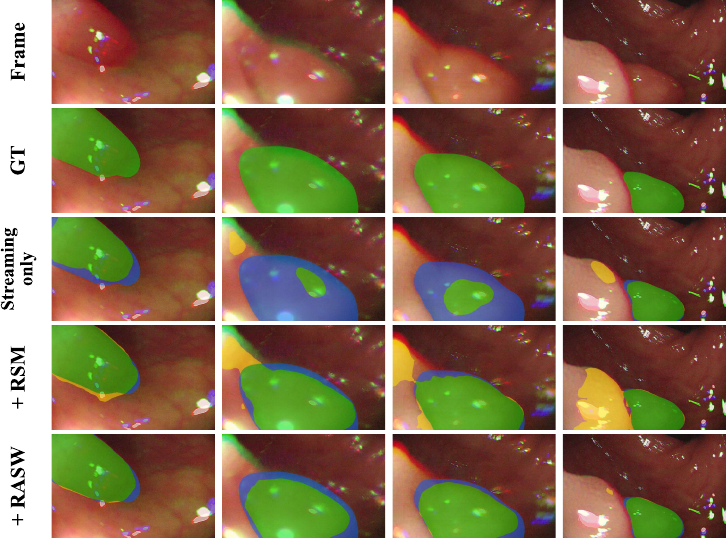}
\caption{\textbf{Visual ablation on a clip (\texttt{case29\_1}) with two blurred frames (middle columns).}
Rows show the input frame, the ground truth, and Streaming only
(\protect\circlednum{1}), RSM with 32 slots (\protect\circlednum{3}), and the
full model (\protect\circlednum{5}) in \cref{tab:abl_step}. Colors as in \cref{fig:qualitative}.}
\label{fig:abl_vis}
\end{figure}

\cref{fig:abl_vis} illustrates this progression on a clip with two
blurred frames. The streaming model without memory misses the polyp on
the blurred frames. RSM restores the mask but carries false positives
from the blurred frames into the next clear frame. RASW removes these
false positives, because its write gates drop on the blurred frames.

\minisection{Number of memory slots.} \Cref{tab:abl_step} compares
RSM with $N\in\{16,32,64\}$ slots while keeping the remaining
architecture fixed and RASW disabled. This isolates the accuracy--state
tradeoff of slot capacity. Reducing the slots to $N=16$ lowers Dice on the unseen sets (76.3\% on Easy-Unseen and 78.3\% on Hard-Unseen),
while $N=64$ brings no consistent gain over $N=32$ (82.1\% and 81.1\%) at a larger state (0.326 vs.\ 0.318~MB), so we use $N=32$.

\begin{table}[pos=!t]
\centering
\caption{Ablation of RASW (Dice, \%). Content: quality-weighted
pooling; State: per-slot acceptance gate. All rows use RSM.
Highl./Blur: the Highlight and Blur subsets of \cref{tab:scenario}.}
\label{tab:abl_write}
\tablesetup
\begin{tabular*}{\linewidth}{@{\extracolsep{\fill}}lcccccc@{}}
\toprule
Write control & Easy-S & Easy-U & Hard-S & Hard-U & Highl. & Blur \\
\midrule
\circlednum{1} None                     & 91.2 & 81.8 & 87.9 & 82.2 & 83.6 & 80.7 \\
\circlednum{2} Content only             & 92.6 & 84.6 & 89.9 & 85.4 & 85.6 & 80.9 \\
\circlednum{3} State only               & 92.2 & 83.8 & 90.0 & 84.8 & 84.0 & 81.6 \\
\circlednum{4} Both (\ourmodel{})       & 92.9 & 85.0 & 90.6 & 85.7 & 86.5 & 83.7 \\
\bottomrule
\end{tabular*}
\end{table}

\minisection{Reliability-aware slot writing.} \cref{tab:abl_write}
switches the two control points of RASW on and off. Compared with
\circlednum{1}, content weighting alone (\circlednum{2}) improves Dice
from 83.6\% to 85.6\% under local highlights, and the per-slot gate
alone (\circlednum{3}) improves it from 80.7\% to 81.6\% under
whole-frame blur. Using both (\circlednum{4}) gives 86.5\% on Highlight and 83.7\% on Blur, because
the content control keeps the clear regions of a partly degraded frame
and the state control protects the slots when no clear region is left.
RSM supplies long-range history and RASW keeps it reliable under local and whole-frame degradation.

\section{Discussion}
\label{sec:discussion}
This section discusses what the results imply for temporal modeling in
VPS, how the data and evaluation protocol may bias them, and where
\ourmodel{} still fails.

\subsection{Significance of the results}
\label{sec:significance}
\minisection{1) Long coverage does not require a growing memory.}
Memory-based video methods usually trade coverage for cost: STM
\citep{oh2019stm} stores every past frame, and XMem
\citep{cheng2022xmem} and SALI \citep{hu2024sali} compress or cap a
bank of past features. Our results suggest that this trade-off is
unnecessary for VPS. \ourmodel{} keeps a constant state of 0.32~MB, whereas
SALI needs 7.46~MB for 35 past frames (\cref{tab:efficiency}), yet
\ourmodel{} achieves higher Dice on all four SUN-SEG test sets
(\cref{tab:easy,tab:hard}). Observation~1 points in the same direction:
a FIFO bank of 256 frames is slower and less accurate than one of 64
frames (\cref{tab:obs1}). What matters is thus not how many frames
are stored, but whether they stay reliable.

\minisection{2) Reliability should be estimated per location, not per
frame.} Existing quality-aware memories decide with one score per frame,
either to select frames \citep{liu2022qdmn,yang2024samurai} or to gate
updates \citep{xiong2026qapa}, and the DMR of CMSA-Net keeps or replaces
whole reference frames. Under spatially non-uniform degradation, such a
decision either writes the degraded regions or discards the clear ones.
The RASW ablation separates the two cases (\cref{tab:abl_write}): the
pixel-level content control raises Dice from 83.6\% to 85.6\% under local
highlights, where part of the frame is still clear, while the per-slot
state control raises Dice from 80.7\% to 81.6\% under whole-frame blur,
where nothing clear is left. This supports Observation~2 with a learned,
rather than an oracle, quality map.

\minisection{3) The gains come mainly from unseen videos.} With
PVTv2-B2, \ourmodel{} improves the Dice of CMSA-Net from 80.3\% to 85.0\%
on Easy-Unseen, from 81.3\% to 85.7\% on Hard-Unseen, and from 80.4\% to
84.3\% on the external CVC-ClinicDB, but only from 92.6\% to 92.9\% on
Easy-Seen. As the Seen subsets share videos with training,
the larger gains on unseen data indicate that RSM and RASW improve
generalization rather than fitting the training videos. The per-video
tests in \cref{sec:sota} support this: the gains are significant on both
unseen sets but not always on the seen sets, where the baselines are
already strong.

\minisection{Clinical relevance.} \ourmodel{} is causal, uses one encoder
pass per frame, and keeps a constant-size state. It runs at 48 FPS (\cref{tab:efficiency}) without buffering
future frames, which fits the online setting of computer-aided
colonoscopy, where a procedure can last many minutes.

\subsection{Data and evaluation considerations}
\label{sec:bias}
\minisection{Dataset scope.} The training and main test data come from
SUN-SEG \citep{sunseg}, so the test videos share the acquisition
conditions of the training videos, and the Seen subsets even share the
same videos. We therefore emphasize the Unseen subsets and evaluate on
CVC-ClinicDB \citep{bernal2015cvc}, which is collected at a different
center, without fine-tuning. Both datasets follow the polyp size and
pathology distributions of their sources, and we do not stratify the
results by histology.

\minisection{Scenario subsets.} The subsets in \cref{sec:scenario} are
defined from images and ground-truth masks only, with fixed quantile
thresholds, so they do not favor any method. Their image statistics
(e.g., Laplacian variance for blur) are proxies for clinical artifacts
and may miss degradations that these statistics do not capture.

\minisection{Synthetic degradation.} \tight The motivating observations and the
streaming training use synthetic blur and highlights, which do not cover
all real artifacts. The gains on the real degradation subsets
(FM, GH, OCC, Blur, and Highlight in \cref{tab:scenario}) suggest that
the learned quality map transfers to real degradation, but this has not
been verified across endoscope systems.

\minisection{Input resolution.} \ourmodel{} processes the full field of
view of every frame. As in the compared methods, frames are resized to
$352\times352$, which keeps the comparison fair and the inference real
time, but may lose detail on very small polyps.

\subsection{Failure case}
\label{sec:failure}
\begin{figure}[pos=!t]
\centering
\includegraphics[width=\columnwidth]{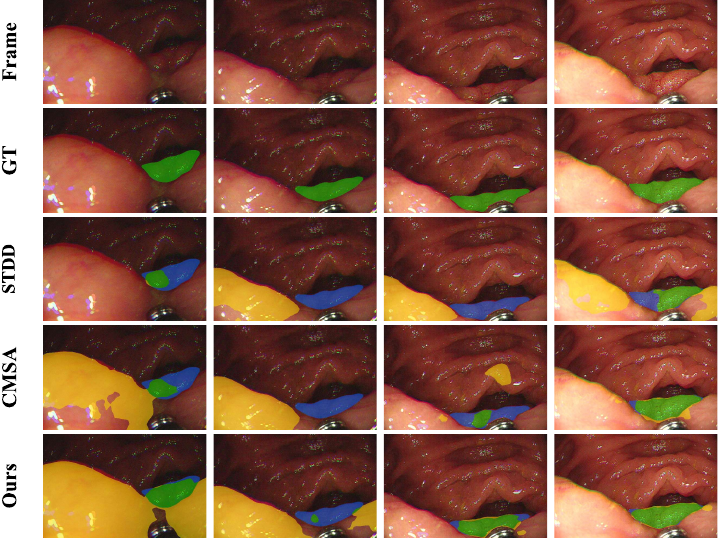}
\caption{\textbf{Failure case of \ourmodel{}.} Long continuous blur in \texttt{case51\_9}
(frames 151--170). Columns show four frames of the clip. Rows show the
input frame, the ground truth, STDDNet, CMSA-Net, and \ourmodel{}. Colors as in \cref{fig:qualitative}.}
\label{fig:failure}
\end{figure}

In \texttt{case51\_9} (\cref{fig:failure}), frames
151--170 are continuously blurred, and \ourmodel{}, STDDNet, and CMSA-Net all
mistake a large blurred fold for the polyp. \ourmodel{} recovers as soon as the
image becomes clear (frame 173) and reaches a Dice of 0.93 at frame 179.

\subsection{Limitations and future work}
\label{sec:future}
Future work will target long continuous blur (\cref{fig:failure}), where little
clear content can be written; we plan to refresh the slots in such intervals
and to test on full-procedure videos.
We will also study other degradations, such as bubbles and smoke, and
extend the memory to other endoscopy tasks. We will further evaluate the
quality map across endoscopes and centers and adapt the slot number to
each video.

\section{Conclusion}
\label{sec:conclusion}
We presented \textbf{\ourmodel{}}, a streaming framework for video polyp
segmentation that gives the causal multi-scale aggregation of CMSA-Net a
reliable long-range history at a constant state. Recursive Slot Memory
(RSM) summarizes the past in a fixed set of slots, and Reliability-Aware
Slot Writing (RASW) uses a pixel-level quality map to decide which
locations and how much of each slot are written, so degraded regions
contribute less while clear ones are kept. On SUN-SEG and CVC-ClinicDB,
\ourmodel{} outperforms existing methods with a state of 0.32~MB that does
not grow with the video, and the gains are largest on unseen videos, indefinable
boundaries, and blurred frames. Recursive, reliability-aware memory is thus a practical basis for online
computer-aided colonoscopy.

\section*{CRediT authorship contribution statement}
\begingroup\raggedright\setlength{\parindent}{0pt}
\textbf{Tong Wang:} Conceptualization, Methodology, Software, Validation,
Formal analysis, Investigation, Visualization, \mbox{Writing -- original draft}.\par
\textbf{Wenbo Peng:} Software, Validation, Investigation, Data curation,
\mbox{Writing -- review \& editing}.\par
\textbf{Yaolei Qi:} Methodology, Formal analysis, Visualization,
\mbox{Writing -- review \& editing}.\par
\textbf{Yuting He:} Methodology, Validation, \mbox{Writing -- review \& editing}.\par
\textbf{Yutong Xie:} Conceptualization, Methodology, Supervision,
\mbox{Writing -- review \& editing}.\par
\textbf{Guanyu Yang:} Conceptualization, Resources, Supervision, Project
administration, Funding acquisition, \mbox{Writing -- review \& editing}.\par
\endgroup

\section*{Declaration of competing interest}
The authors declare that they have no known competing financial interests
or personal relationships that could have appeared to influence the work
reported in this paper.

\section*{Data availability}
SUN-SEG \citep{sunseg} and CVC-ClinicDB \citep{bernal2015cvc} are
publicly available. The code will be released at
\url{https://github.com/wangtong627/STREAM}.

\section*{Declaration of generative AI and AI-assisted technologies in the manuscript preparation process}
During the preparation of this work the authors used
ChatGPT (OpenAI) to polish the language and
improve the readability of the manuscript. After using this tool, the
authors reviewed and edited the content as needed and take full
responsibility for the content of the published article.

\section*{Acknowledgements}
Tong Wang, Wenbo Peng, Yaolei Qi, and Guanyu Yang are supported by the
National Natural Science Foundation of China (T2541064, 82441021) and the
Jiangsu Province Cutting-edge Technology R\&D Program (BF2025628). This
work is also supported by the Big Data Computing Center of Southeast
University.

\bibliographystyle{cas-model2-names}
\bibliography{references}

\end{document}